\pdfoutput=1
\documentclass[sigconf]{acmart}
\usepackage[ruled, linesnumbered, longend, nofillcomment]{algorithm2e}
\usepackage{listings}
\usepackage{multirow}
\usepackage{makecell}
\usepackage[para,online,flushleft]{threeparttable}
\usepackage{enumerate}
\usepackage{enumitem}
\usepackage{balance}
\AtBeginDocument{%
  \providecommand\BibTeX{{%
    \normalfont B\kern-0.5em{\scshape i\kern-0.25em b}\kern-0.8em\TeX}}}

\setcopyright{rightsretained}
\copyrightyear{2024}
\acmYear{2024}
\acmDOI{10.1145/3637528.3671620}

\acmConference[KDD '24]{Proceedings of the 30th ACM SIGKDD Conference on Knowledge Discovery and Data Mining}{August 25--29, 2024}{Barcelona, Spain}
\acmBooktitle{Proceedings of the 30th ACM SIGKDD Conference on Knowledge Discovery and Data Mining (KDD '24), August 25--29, 2024, Barcelona, Spain}
\acmISBN{979-8-4007-0490-1/24/08}

\newcommand{\model}{\textsc{AutoWebGLM}\xspace}
\newcommand{\chatgpt}{GPT-3.5-Turbo}
\newcommand{\gptfour}{GPT-4}
\newcommand{\benchmark}{AutoWebBench\xspace}

\newcommand{\vpara}[1]{\vspace{0.04in}\noindent\textbf{#1}\xspace}

\newcommand{\hide}[1]{}

\makeatletter
\gdef\@copyrightpermission{
  \begin{minipage}{0.3\columnwidth}
   \href{https://creativecommons.org/licenses/by/4.0/}{\includegraphics[width=0.90\textwidth]{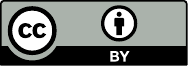}}
  \end{minipage}\hfill
  \begin{minipage}{0.7\columnwidth}
   \href{https://creativecommons.org/licenses/by/4.0/}{This work is licensed under a Creative Commons Attribution International 4.0 License.}
  \end{minipage}
  \vspace{5pt}
}
\makeatother

\begin{document}


\title{AutoWebGLM: A Large Language Model-based Web Navigating Agent}

\author{Hanyu Lai}
\email{laihy23@mails.tsinghua.edu.cn}
\affiliation{
  \institution{Tsinghua University}
\city{Beijing}
\country{China}
}
\authornote{HL, XL, and ILI contributed equally to this research.}
\authornote{Work done while these authors interned at Zhipu AI.}

\author{Xiao Liu}
\authornotemark[1]
\email{shawliu9@gmail.com}
\affiliation{
  \institution{Tsinghua University \& Zhipu AI}
   \city{Beijing}
   \country{China}
}

\author{Iat Long Iong}
\authornotemark[1]
\authornotemark[2]
\email{rongyl20@mails.tsinghua.edu.cn}
\affiliation{
  \institution{Tsinghua University}
\city{Beijing}
\country{China}
}

\author{Shuntian Yao}
\authornotemark[2]
\email{yaoshuntian@bupt.edu.cn}
\affiliation{
  \institution{Beijing U. of Posts and Telecoms}
\city{Beijing}
\country{China}
}

\author{Yuxuan Chen}
\authornotemark[2]
\email{chenyuxu21@mails.tsinghua.edu.cn}
\affiliation{
  \institution{Tsinghua University}
\city{Beijing}
\country{China}
}

\author{Pengbo Shen}
\authornotemark[2]
\email{pengbo.shen@outlook.com}
\affiliation{
  \institution{U. of Chinese Academy of Sciences}
\city{Beijing}
\country{China}
}

\author{Hao Yu}
\authornotemark[2]
\email{longinyh@gmail.com}
\affiliation{
  \institution{Tsinghua University}
\city{Beijing}
\country{China}
}

\author{Hanchen Zhang}
\authornotemark[2]
\email{hc-zhang22@mails.tsinghua.edu.cn}
\affiliation{
  \institution{Tsinghua University}
\city{Beijing}
\country{China}
}

\author{Xiaohan Zhang}
\email{xiaohan.zhang@zhipuai.cn}
\affiliation{
  \institution{Zhipu AI}
\city{Beijing}
\country{China}
}

\author{Yuxiao Dong}
\email{yuxiaod@tsinghua.edu.cn}
\affiliation{
  \institution{Tsinghua University}
\city{Beijing}
\country{China}
}
\authornote{Corresponding Authors: YD and JT.}

\author{Jie Tang}
\authornotemark[3]
\email{jietang@tsinghua.edu.cn}
\affiliation{
  \institution{Tsinghua University}
\city{Beijing}
\country{China}
}

\renewcommand{\shortauthors}{Hanyu Lai et al.}

\begin{abstract}
Large language models (LLMs) have fueled many intelligent web agents, but most existing ones perform far from satisfying in real-world web navigation tasks due to three factors: (1) the complexity of HTML text data (2) versatility of actions on webpages, and (3) task difficulty due to the open-domain nature of the web. 
In light of these challenges, we develop the open \model based on ChatGLM3-6B. 
\model can serve as a powerful automated web navigation agent that outperform GPT-4.  
Inspired by human browsing patterns, we first design an HTML simplification algorithm to represent webpages with vital information preserved succinctly. 
We then employ a hybrid human-AI method to build web browsing data for curriculum training.
Finally, we bootstrap the model by reinforcement learning and rejection sampling to further facilitate webpage comprehension, browser operations, and efficient task decomposition by itself.
For comprehensive evaluation, we establish a bilingual benchmark---\benchmark---for real-world web navigation tasks. 
We evaluate \model~across diverse web navigation benchmarks, demonstrating its potential to tackle challenging tasks in real environments. 
Related code, model, and data are released at \url{https://github.com/THUDM/AutoWebGLM}.
\end{abstract}

\ccsdesc[500]{Computing methodologies~Intelligent agents}

\keywords{ChatGLM, Large Language Model, LLM Agent, Web Agent, Reinforcement Learning, Rejection Sampling Finetuning}



\maketitle

\begin{figure}[h]
 \vspace{-5mm}
  \includegraphics[width=\linewidth]{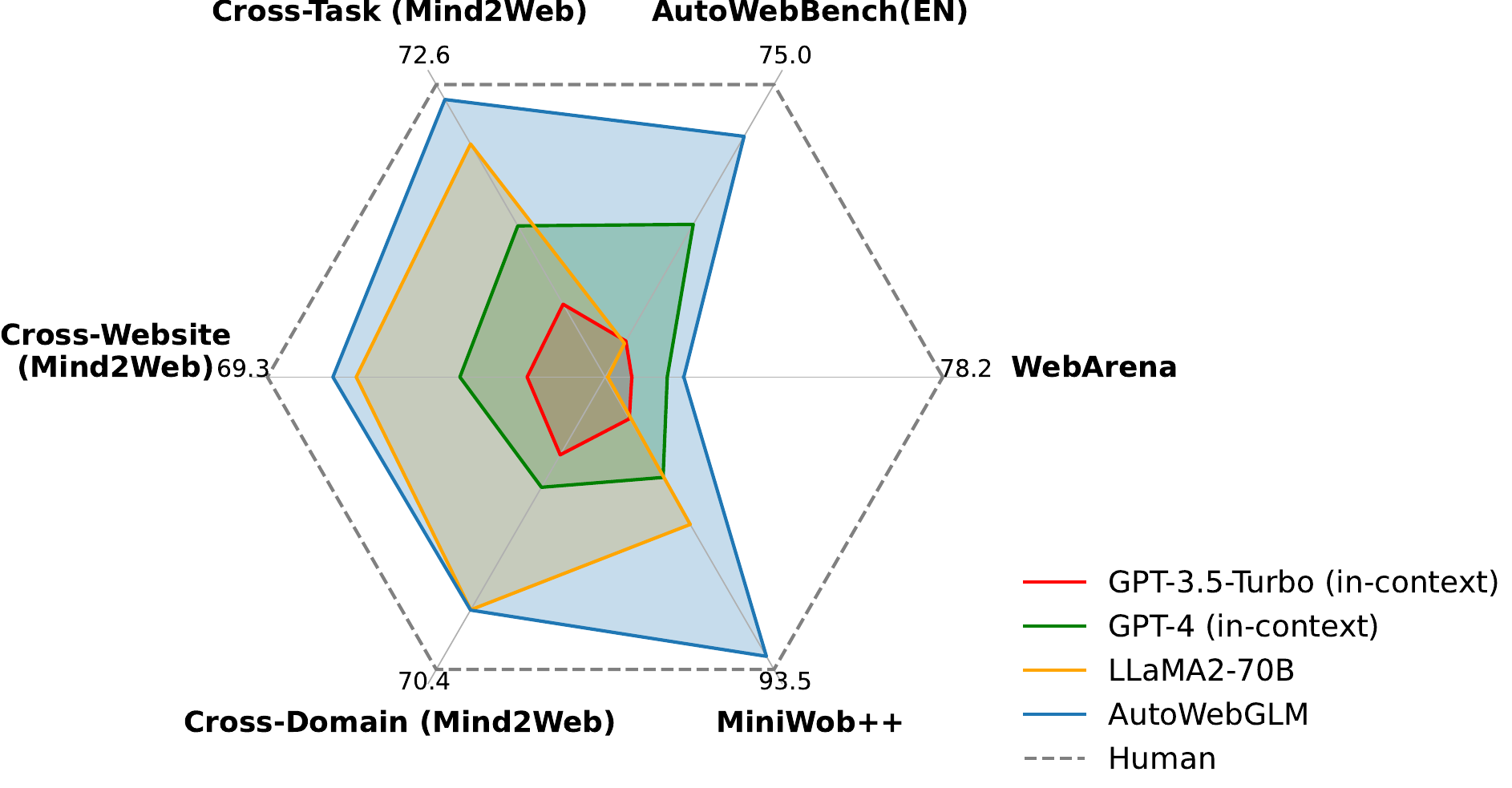}
  \vspace{-7mm}
  \caption{The performance of \model on various web browsing tasks in comparison with GPT-4 and open LLMs.}
  \vspace{-5mm}
  \label{fig:overall_result}
\end{figure}

\begin{figure*}[t!]
  \includegraphics[width=.9\linewidth]{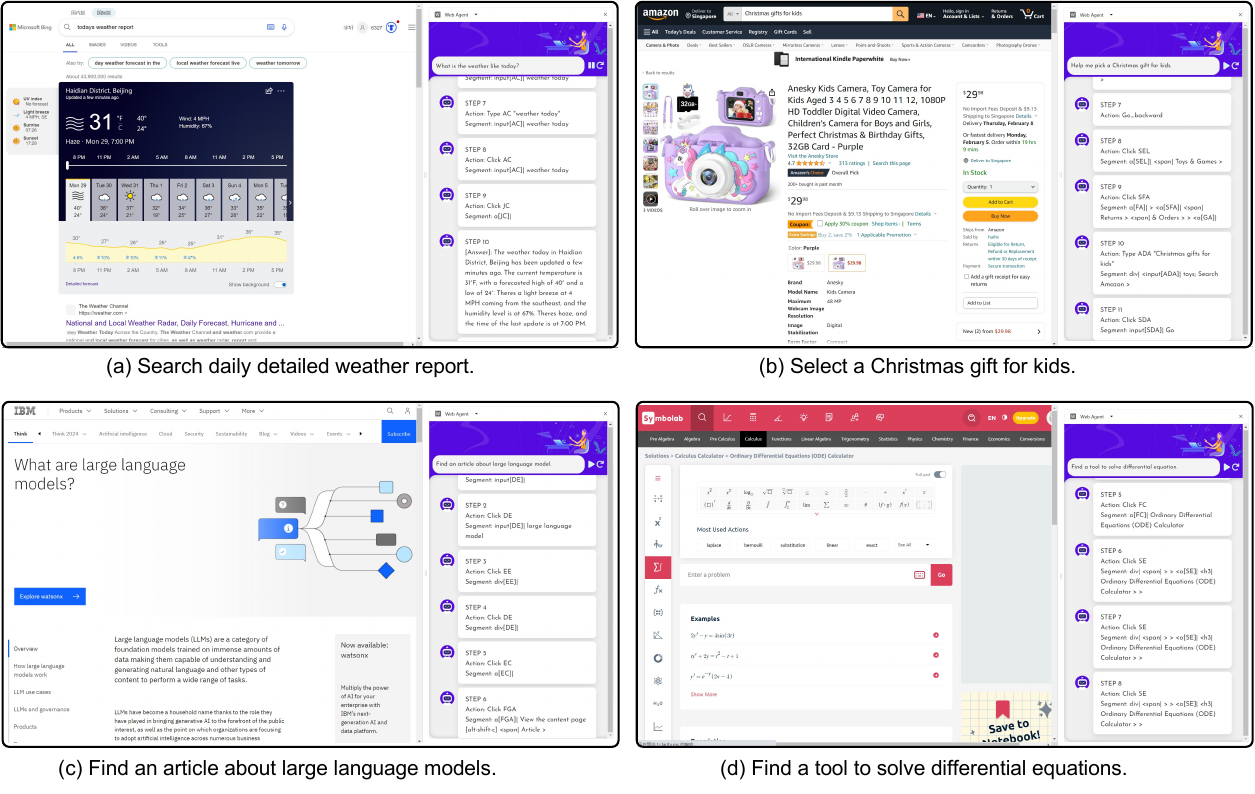}
    \vspace{-3mm}
  \caption{Examples of \model's execution on four user tasks.}
  \vspace{-3mm}
  \label{fig:demo}
\end{figure*}

\section{Introduction}
The concept of autonomous digital agents as helpful assistants is an enticing prospect. 
Enhanced by LLMs' formidable comprehension and response capabilities~\cite{achiam2023gpt, touvron2023llama, touvron2023llama2, zeng2022glm, zhang2022opt, scao2022bloom}, we can envision various scenarios previously unimaginable. 
For instance, an LLM-based agent could support a daily routine that summarizes the online news from the open web for us. 
This integration of LLMs into everyday tasks heralds a significant shift in how we interact with machines, optimizing our efficiency and redefining the boundaries of machine-assisted productivity~\cite{xi2023rise, wang2023survey}. 

Tremendous efforts have been underway to construct auto web agents. 
One is AutoGPT, a popular open-source project that utilizes ChatGPT~\cite{chagpt} to integrate LLMs with predetermined tools such as web and local file browsing.
Meanwhile, the development of agent-centric LLMs has gained significant momentum~\cite{yao2022react, press2022measuring, wang2023plan,hong2023cogagent}.
Nevertheless, the majority of existing web agents are to date severely restricted in terms of practical applications, predominantly due to the following challenges: 

\begin{itemize}[leftmargin=*,itemsep=0pt,parsep=0.2em,topsep=0.2em,partopsep=0.0em]
\item A universal action space covering all necessary task executions across various websites is absent.
\item The diversity and complexity of webpages and their tendentious verbosity pose a significant challenge for LLMs to comprehend the content and carry out correct operations accurately.
\item Existing agents notably lack the capability for correct inference and self-checking on web tasks.
Once caught in an erroneous loop, they struggle to rectify the issue promptly.
\end{itemize}

In this work, we introduce \model for building webpage navigation agents. 
It is built upon the open ChatGLM3-6B model~\cite{zeng2022glm}. 
First, we propose various efficient data strategies to support the swift construction of a sizeable, reliable training dataset while state-of-the-art models cannot reliably complete data annotation tasks~\cite{zhou2023webarena}. 
Furthermore, by leveraging supervised~\cite{ouyang2022training} and reinforcement learning methods~\cite{rafailov2023direct}, we train \model on top of the collected web agent dataset to achieve performance superiority on general webpage browsing tasks. 
A step further, we employ rejection sampling finetuning (RFT) ~\cite{touvron2023llama2} for lifelong learning in specific web environments, enabling the agent to excel in a particular domain.

We develop and deploy a Chrome extension
based on \model (See Figure~\ref{fig:demo} for examples).  
Throughout our experiments, it can reason and perform operations on various websites to complete user tasks accurately, making it practically applicable to real-world services. 
In addition, 
we construct the first bilingual (English and Chinese) webpage browsing evaluation dataset to build \benchmark, given that websites from different regions have substantial stylistic variations. 
In conclusion, we make the following contributions in this paper:

\begin{itemize}[leftmargin=*,itemsep=0pt,parsep=0.2em,topsep=0.2em,partopsep=0.0em]
    \item We design and develop the \model agent for effectively completing web browsing tasks through curriculum learning, bootstrapped by self-sampling reinforcement learning, and RFT in the web browsing environment.     
    \item We construct a real webpage browsing operation dataset of approximately 10,000 traces using model-assisted and manual methods, including the bilingual (English and Chinese) web browsing benchmark \benchmark.
    \item We perform experiments to demonstrate that \model with six billion parameters achieves performance comparable to the most advanced LLM-based agents, and more importantly, it reaches a practically usable level for real-world web tasks. 
\end{itemize}

\section{Related Work}
Constructing a comprehensive web browsing agent is a complex task that involves various modules, such as a language model for decision-making and an HTML parser for environment observation. Furthermore, it is essential to have appropriate web browsing evaluation criteria when creating an effective web browsing agent. In this section, we will discuss the works related to these aspects.

\vpara{Language Models (LMs).} Large language models (LLMs)~\cite{zhao2023survey}, such as GPT-4~\cite{achiam2023gpt}, Claude-2~\cite{claude2}, LLaMA-2~\cite{touvron2023llama}, ChatGLM~\cite{zeng2022glm,du2022glm}, OPT~\cite{zhang2022opt}, and BLOOM~\cite{scao2022bloom}, have accumulated extensive knowledge in various natural language processing tasks. However, due to the high cost of deploying such large language models, smaller models with lower costs and comparable capabilities are usually preferred. Many open-source projects, such as LLaMA-2-7B~\cite{touvron2023llama} and ChatGLM3-6B~\cite{zeng2022glm}, have demonstrated strong performance to large language models in some domains. 

\vpara{Benchmarks for Web Navigation.} The primary web browsing evaluation datasets provide a variety of evaluation metrics. MiniWoB++~\cite{humphreys2022data} provides several simulated web environments, with tasks primarily to evaluate the model's ability to interact with webpage components. However, with the increasing demand for complex web operation capabilities, Mind2Web~\cite{deng2023mind2web} and WebArena~\cite{zhou2023webarena} have been created. 
Mind2Web is an offline evaluation set for complex web browsing that provides several metrics. The evaluation method is straightforward and commonly used for model evaluations. 
In contrast, the WebArena benchmark, based on real websites, creates multiple virtual environments and uses various evaluation methods to assess the task completion rate, making it more suitable for real-world task completion evaluation.

\vpara{Agents for Web Automation.} 
Previous work such as WebGPT~\cite{nakano2021webgpt} and WebGLM~\cite{liu2023webglm} combined LLMs with web environments. However, their primary application was question-answering (QA) tasks~\cite{rajpurkar2016squad, nguyen2016ms,berant2013semantic,kwiatkowski2019natural}, utilizing internet resources to answer user queries.
Recent works~\cite{mishra2019cc, hong2023cogagent, cheng2024seeclick, xu2023lemur} focus more on executing complex operations or interactive tasks.
A fundamental aspect of web browsing tasks involves a comprehensive understanding of HTML. StructGPT~\cite{jiang2023structgpt} explores methodologies to enhance the zero-shot reasoning ability of LLMs in handling structured data.
Specifically, MindAct~\cite{deng2023mind2web} works by filtering webpage elements and selecting the element through multiple rounds of multiple-choice questions. It often requires more than ten model calls to complete a single web operation, which could be more efficient.
On the other hand, WebAgent~\cite{gur2023real} uses HTML-T5 to process the observation space's content, including HTML, previous operations, and user instructions.
It uses the Flan-U-Plam model to generate code to control webpages, exhibiting excellent web browsing performance. However, it faces deployment challenges due to the size of the Flan-U-Plam model, which is 540B scale.
\model, based solely on a single ChatGLM3-6B, has a robust web browsing capability comparable to WebAgent, demonstrating high value for practical deployment.

\vpara{Prompt-based Data Construction Methods.} 
Constructing data through prompts has recently gained significant traction~\cite{wang2022self, honovich2022unnatural, peng2023instruction, mukherjee2023orca, cheng2023black}. 
This approach leverages language models to generate synthetic data for training. 
A notable example is Evol-Instruct~\cite{xu2023wizardlm, luo2023wizardcoder}, inspired by the theory of evolution, demonstrating the effectiveness of using LLMs to generate diverse and complex instructions for various tasks.
Additionally, some researchers explore the potential of generating data in a zero-shot setting, where the model produces data for tasks it has yet to be explicitly trained on~\cite{meng2022generating}, highlighting the versatility of prompt-based data construction. 
These methodologies rapidly evolve, offering a promising avenue for data generation in various domains, especially where traditional data collection methods could be more practical and sufficient..

\vpara{Rejection Sampling Finetuning.}
The methodology of Rejection Sampling Finetuning (RFT)~\cite{yuan2023scaling} employs a supervised learning model to generate and collect accurate reasoning paths, subsequently utilized as an augmented finetuning dataset. Using RFT to expand the dataset with diverse reasoning paths can boost the mathematical performance of LLMs. Our experiments show that RFT can also be effectively implemented in web page browsing tasks, significantly increasing professional capabilities within specific environments.

\begin{figure*}[t]
  \centering
  \includegraphics[width=.95\linewidth]{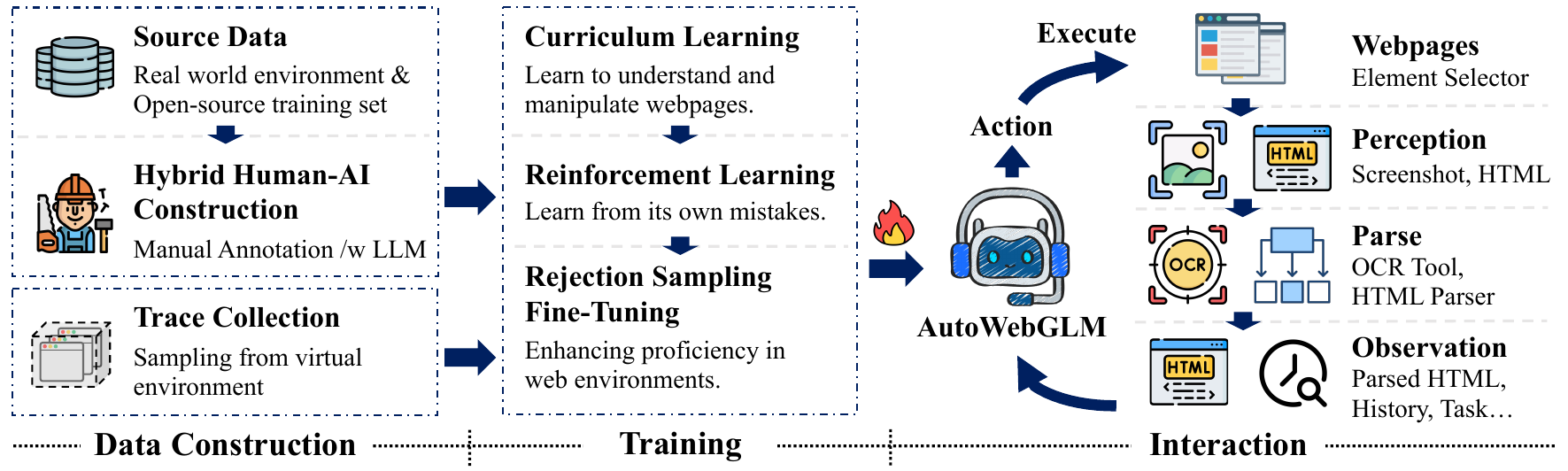}
  \caption{The System Architecture of \model. \textmd{Our system comprises two key components: interaction framework and LM agent.
  The LM agent learns from data procured from diverse sources. It further employs RL and RFT to bootstrap itself, thus enhancing web browsing capabilities.
  The interaction framework uses various web processing modules to organize concise HTML and other information for the LM agent to make decisions that are then executed by an automated browsing program.
  }}
  \label{fig:framework}
\end{figure*}

\section{\model as A Web Agent}
\subsection{Problem Setup}
We consider web browsing tasks as a sequential decision-making process. 
The state, denoted as $S$, includes the current page status, such as HTML, URL, and Window Position. 
The action set $A$ contains all potential browsing operations, including click, type, scroll, etc (See complete operations in Table~\ref{tab:action_space}). 
\[
S = \{ \text{HTML}, \text{URL}, \text{Window Position} \},  A = \{ \text{click}, \text{scroll}, \text{type}, \ldots \}
\]
The state's transition is determined by the webpage's current state and the agent's output action. 
During the decision-making process, the function \( \phi \) updates the historical information based on the previous history $H_{t-1}$, the most recent action $A_{t-1}$, and the current state $S_t$.
\[
H_t = \phi(H_{t-1}, A_{t-1}, S_t)
\]
The policy \( \pi \) is the process for the agent to choose actions based on the current state and the history. 
A complete decision process starts from the initial state $S_0$ and history $H_0$, iterating through the policy \( \pi \) and transition function $T$. 
This iteration ceases when the action $A_t$ is \textit{finish} or reaches the maximum length.
\[
(S_{t+1}, H_{t+1}) = (T(S_t, A_t), \phi(H_t, A_t, S_{t+1}))
\]
\[
A_t = \pi(S_t \mid H_t)
\]
\[
S_{t+1} = T(S_t, A_t)
\]

\begin{table}[htbp]
    \small
    \caption{Action space for \model~to interact through.}
    \vspace{-4mm}
    \renewcommand\tabcolsep{2pt}
    \label{tab:action_space}
    \begin{tabular}{ll}
        \toprule
        Instruction & Description\\
        \midrule
        \texttt{click(id)} & Click at an element\\
        \texttt{hover(id)} & Hover on an element\\
        \texttt{select(id, option)} & Select option in an element\\
        \texttt{type\_string(id, text, enter)} & Type to an element\\
        \midrule
        \texttt{scroll\_page(direction)} & Scroll up or down of the page\\
        \texttt{go(direction)} & Go forward or backward of the page\\
        \texttt{jump\_to(url, newtab)} & Jump to URL\\
        \texttt{switch\_tab(id)} & Switch to i-th tab\\
        \midrule
        \texttt{user\_input(message)} & Notify user to interact\\
        \texttt{finish(answer)} & Stop with answer\\
        \bottomrule
    \end{tabular}
    \vspace{-5mm}
\end{table}

\subsection{The \model Framework}\label{framework}
As depicted in Figure~\ref{fig:framework}, we process information through HTML simplification and OCR (Optical Character Recognition) modules, yielding a simplified HTML representation after acquiring HTML and webpage screenshots. 
With attributes facilitating operability judgment, we mark operable elements for agent interaction. 
The OCR module is for notating text elements during image parsing.

Agents initiate action prediction by combining this representation with other observational data. 
Upon outputting action, the automated web program is employed for action execution; this iterative cycle persists until task termination. 
\model~enhances interactive capacity and webpage navigation precision by amalgamating these components into a singular framework.

A comprehensive, precise observation and action space is vital for constructing a robust web browsing framework. 
These spaces standardize the conversion of varied data sources into a uniform format.
We discuss our designs in the following:

\subsubsection{Observation space}
We suggest using a unified observation space to enhance the model's webpage comprehension and operation level. The observation space should provide information as close as possible to what the browser's graphical interface can provide, thus maximizing the upper bound of the agent's capabilities.
We identify four critical indicators for web browsing tasks: task description, simplified HTML, current location, and past operation records. The HTML provides the model with structural and content information about the page, while the current location information helps the model understand its position within the webpage. The record of past operations provides the model with historical context, which helps to generate more consistent subsequent operations.
By incorporating these elements into the observation space, we strive to construct a more resilient and practical model that can handle the intricacy and variability inherent in web browsing tasks. The following are detailed illustrations of the observation space components.

\begin{algorithm}
    \DontPrintSemicolon
    \SetAlgoLined
    \SetNoFillComment

    \caption{HTML Pruner}

    \KwData{tree $tree$, kept elements $kept$, recursion count $rcc$, max depth $d$, max children $mc$, max sibling $ms$}
    \KwResult{pruned tree $tree$}

    \SetKwFunction{Fans}{getAnscendants}
    \SetKwFunction{Fdes}{getDescendants}
    \SetKwFunction{Fsib}{getSiblings}
    \SetKwFunction{Flen}{len}
    \SetKwFunction{Fupt}{update}
    \SetKwFunction{Fapp}{append}
    \SetKwFunction{Frev}{reversed}
    \SetKwFunction{Frem}{remove}
    \SetKwFunction{Fbrac}{}
    \SetKwProg{Fn}{Function}{:}{}
    \SetKw{KwIn}{in}
    \SetKw{KwNot}{not}
    \SetKw{KwAnd}{and}
    \SetKw{KwOr}{or}

    $nodes\gets []$\;
    \For{$t\gets0$ \KwTo $rcc$}{
        \For{$id$ \KwIn $kept$}{
            $node\gets tree.element$ with $id$\;
            $nodes.\Fapp{node}$\;
            $nodes.\Fapp{\Fans{node, d}}$\;
            $nodes.\Fapp{\Fdes{node, d, mc}}$\;
            $nodes.\Fapp{\Fsib{node, ms}}$\;
        }
        $d,mc,ms\gets \Fupt{d,mc,ms}$\tcp{make them smaller}
    }

    \For{$node$ \KwIn $\Frev{tree}$}{
        \If{\KwNot $node$ \KwIn $nodes$ \KwOr \KwNot \Fbrac{$node$ has text or attrib \KwOr $\Flen{node.children}>1$ \KwOr $node$ is root}}{
            $tree.\Frem{element}$\;
        }
    }
    
\end{algorithm}

\vpara{HTML.} The HTML webpages are vast and complex, so it is necessary to simplify them before inputting them into the model. 
The simplification process aims to extract essential information while eliminating redundant or disruptive elements that could hinder the model's understanding. Throughout this process, the HTML's basic structure and significant content information must be retained to enable the model to comprehend and utilize this information for effective web browsing.
HTML Pruner can efficiently convert a tree of elements into a concise representation.
We can use the processing techniques to streamline the original HTML format into a more understandable structure for the model to interpret and manage, improving model effectiveness in web browsing tasks.

\vpara{Current Position.} Based on our observation of the model's interaction with the environment, agents could perform better when provided with window position and page size.
The agent uses the page scroll position to understand the content of the currently visible area and the page height information to comprehend the scale of the entire page, providing a spatial context for the model. 

\vpara{Previous actions.} The best solution to inform the agent of past operations is explicitly providing it. This approach helps the agent understand its past behaviors. It prevents the agent from getting stuck in an ineffective loop of repeating the same actions due to operational failures, improving its ability to adapt to the complexities and dynamics of web browsing tasks by preventing the recurrence of unsuccessful operations.

\subsubsection{Action space}
As the approach of this work is to build a language model-based web browsing agent, we focus on operational possibilities when constructing the action space.
On an extensive summary of experiences in the real task execution process, we define a complete and self-consistent action space (in Table~\ref{tab:action_space}) for the language model to act in the web browsing world.
We design our prompt input in Section~\ref{appendix:input_prompt}.
\section{Building \model}

\begin{figure*}[t]
    \centering
    \includegraphics[width=.95\linewidth]{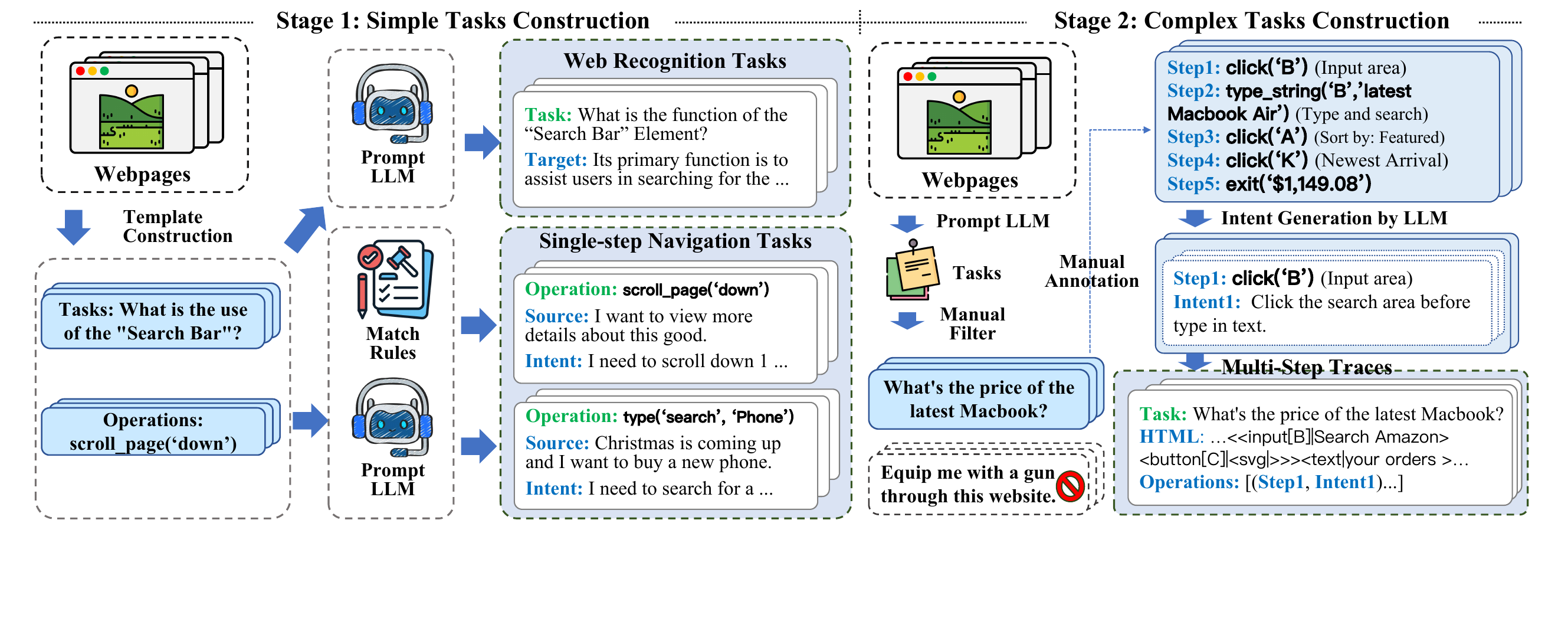}
    \caption{Data Construction. \textmd{Data construction is divided into two main stages; the first stage is webpage recognition tasks and simple tasks operation construction, and the second stage is complex tasks construction.}}
    \label{fig:data_construction}
\end{figure*}

In this section, we detail the construction of a web browsing agent. Given the high costs associated with manual data construction and the inadequacy of current LLMs for automated data generation, we employed a Human-AI hybrid data construction method to efficiently produce large volumes of training data at a reduced cost. Additionally, we implemented a multi-stage training approach, rather than relying solely on imitation learning, to enhance our model's general and specialized web browsing capabilities.

\subsection{Data Construction}\label{data_preparation}
Considering the scarcity of high-quality, complex web browsing data produced by actual users, we aim to create a training dataset. However, the dataset construction presents several challenges:
\begin{itemize}[leftmargin=*,itemsep=0pt,parsep=0.2em,topsep=0.2em,partopsep=0.0em]
    \item Task Collection: A significant hurdle is acquiring diverse, real-user task queries across various websites.

\item Privacy and Security: Privacy and security limitations hinder the direct acquisition of user browser operation sequences. It is also challenging to rule out redundant or incorrect operations not pertinent to task completion and to confirm user task completion.

\item Objective Annotation: The labor-intensive nature of collecting user objectives for each operational step makes it impractical in real-world data-gathering scenarios.

\item Model Limitations: Current models cannot process complex user queries across different websites, thus eliminating the chance of using purely automated methods for accurate browsing trajectory collection in real and complex application contexts.
\end{itemize}

As illustrated in Figure~\ref{fig:data_construction}, we suggest a hybrid human-AI Data Construction method to create our training data in response to these challenges.
After careful consideration, we categorize our data into two types for construction:

\subsubsection{Web Recognition \& Simple Task Operation Construction}~\label{single_step}
For web browsing tasks, efficient and accurate understanding and manipulation of webpages become vital challenges in model development due to the diversity of user behaviors and the complexity of web content.
This section illustrates our construction method for web recognition and simple task operation to train models to recognize webpage structures and perform basic operations accurately.

\vpara{Web Recognition.}
The main objective of Web Recognition includes understanding particular HTML formats, identifying different types of web elements (such as text boxes, buttons, images, etc.), and understanding the role of these elements in user interaction. We propose the following construction approach based on the above practical challenges.

We initiate our process by collecting URLs from Chinese and English mainstream websites listed on Similarweb\footnote{https://www.similarweb.com/top-websites}. In the data processing stage, we use our HTML parser to identify operable components in each webpage and record essential information such as component position and size. We then generate a simplified HTML by rearranging and simplifying the component tree (see details in Section~\ref{framework}).

We design tasks such as website and component function descriptions to aid model recognition of webpage structures and interactive components' functions. 
For each task, we develop a series of natural language questions to serve as the source field in our data. \chatgpt~is utilized to generate multiple formulations for each question, thereby diversifying the question formation.

For the target, we leverage \chatgpt~to generate the response. 
We supply a simplified HTML with the pertinent question in the prompt and impose a limit on the response length, thereby obtaining our target.

\vpara{Simple Task Operation.}
The main objective of the Simple Task Operation dataset is to train models to perform single-step web operations. This involves executing basic functionalities on web pages, such as clicking links, filling out forms, or navigating to specific sections. 
To build our data, we collect various websites in the same way as Web Recognition.
Then, we construct a split for each operation type to ensure that our dataset covers all the requirements for simple task operations. We adjust the data size for each split based on the frequency of each operation in practice.

Our key to constructing the dataset is by rules instead of model generation.
We try \chatgpt~for tasks, intent, and operation generation and Selenium~\footnote{https://www.selenium.dev} to validate the executability of the generated results.
However, it has obvious drawbacks: The model cannot reach an acceptable accuracy in the operation to fulfill the task, and the correctness of the model-generated operations is hard to judge.
To address the above issues, we endeavor to approach from a novel perspective.
We identify various actionable elements within the webpage, assembling them into web operations. 
Then, we use \chatgpt~to produce the corresponding tasks and operational intents for these actions.
For operation types with relatively fixed behaviors, such as Scroll and Jump\_to, we directly generate their corresponding tasks with templates; for flexible and feature-rich operations, such as Click and Type, we use \chatgpt~to help complete the construction.
This approach ensures the instructions' executability and provides the operation tasks' richness.

\subsubsection{Complex Task Operation Construction}~\label{multi_step}
We developed a dataset for complex web tasks to enable the model to make plans and reason in the web browsing scenario. Each sample in the dataset consists of a real-world complex web browsing task, the sequence of operations to complete the task, and the intent of each step.

We first designed 50 complex tasks for each website using the prompting technique referring to Evol-Instruct~\cite{xu2023wizardlm}, from which about 20 feasible tasks were manually selected and labeled.
For operation sequence, due to the high complexity of the tasks, even the most advanced LLMs cannot complete the task with satisfactory accuracy. 
Therefore, we leveraged manual annotations to capture web task executions via a browser plugin that records actions during website tasks.
Chain-of-thought \cite{wei2022chain} reasoning has been proven to improve task comprehension and model performance \cite{kojima2022large, wang2022self} significantly. 
However, leveraging human annotators to document their intent and reasoning during web browsing is inefficient. 
To improve the CoT construction process, we used \gptfour~as the operational intent predictor.
Our first approach of iterative step-by-step creation proved to generate weak operational links and incurred high API costs due to data construction. 
To address this, we employed a global thought chain prompting method, where all operations and critical HTML segments are inputted into a trace. Then, we prompted \gptfour~to output intentions for each step. This method improves the accuracy and cohesion of each step, thus forming highly relevant, consistent thought chains.

\begin{figure}[htbp]
    \centering
    \vspace{-2mm}
    \includegraphics[width=0.8\linewidth]{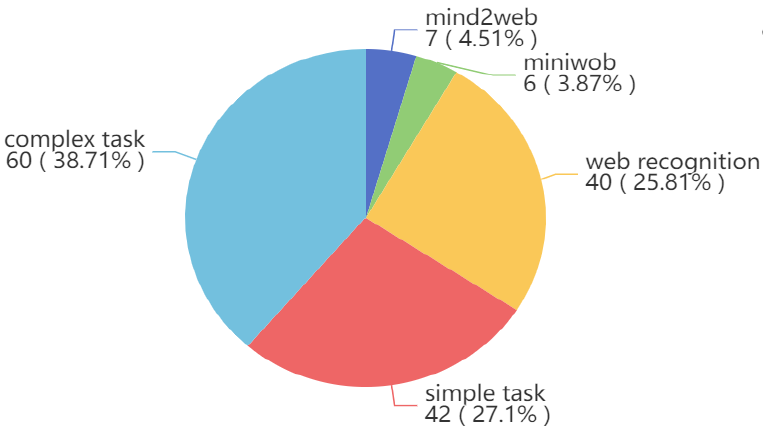}
    \vspace{-3mm}
    \caption{Dataset Proportion. \textmd{Piechart of the distribution of splits within our training data.}}
    \vspace{-5mm}
    \label{fig:data_proportion}
\end{figure}

After construction, we merge our data with the training set from Mind2Web and MiniWob++ to form our final training dataset. The proportion of each split is in Figure~\ref{fig:data_proportion}.

\begin{figure*}[t]
    \centering
    \includegraphics[width=0.95\linewidth]{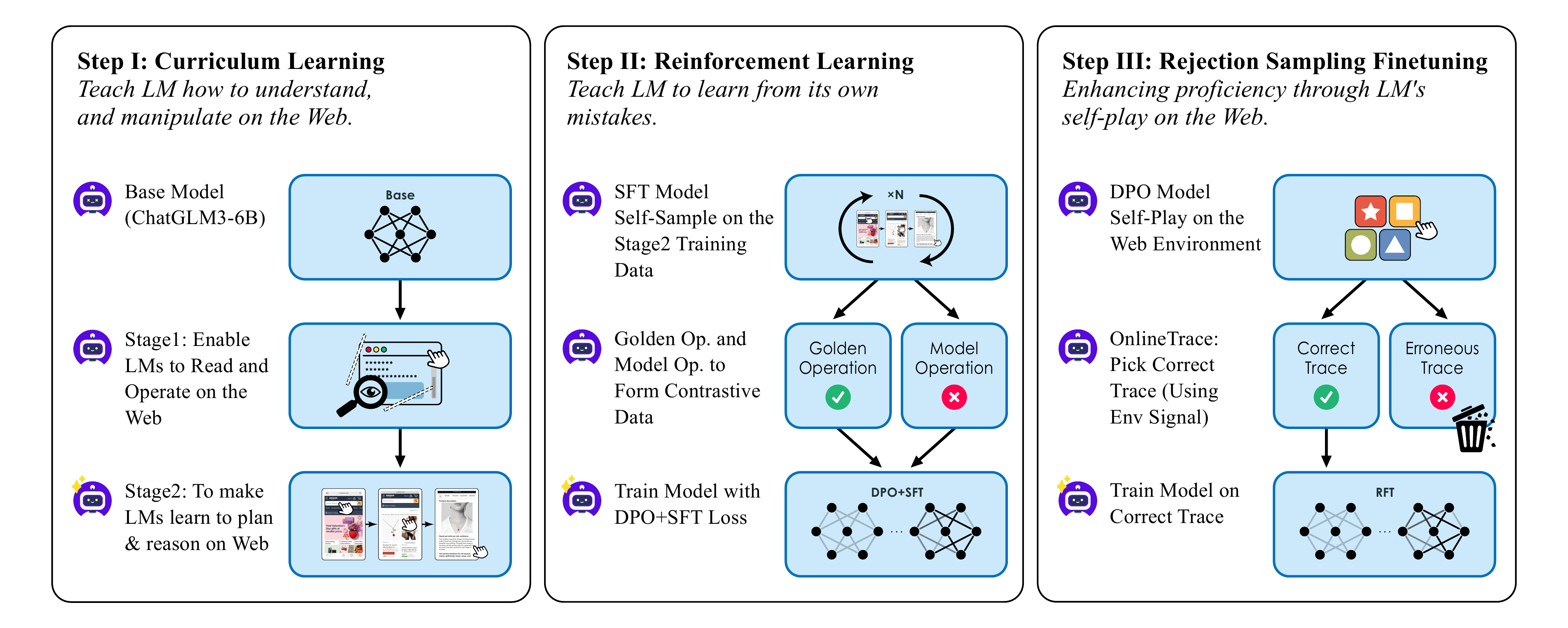}
    \caption{The Training Procedure. \textmd{First, the model learns webpage interpretation and operation via curriculum learning. Next, it self-samples training data, learning from its mistakes. Finally, it self-plays in the environment, becoming a domain expert.}}
    \label{fig:training}
\end{figure*}

\subsection{Training}\label{training}
We train the model through three steps illustrated in Figure~\ref{fig:training}.
\subsubsection{Step 1: Curriculum Learning}\label{sft}
The first one is Supervised Fine-Tuning (SFT). We utilize data in Section~\ref{data_preparation} for training
\begin{equation}
\mathcal{L_{SFT}}(\pi_\theta)=-\mathbb{E}_{(x,y)\thicksim\mathcal{D}}\left[\log \pi_\theta(y\mid x)\right]
\end{equation}

This approach enhances the model's comprehension of webpages and its capability as an agent to perform operations within the environments. 
Significantly, we use curriculum learning (CL), which mimics the human learning process, advocating for models to start learning from easy samples and gradually advance to complex ones.
It has been demonstrated in prior works\cite{bengio2009curriculum, wang2021survey} to improve model capabilities substantially. 

\vpara{Enabling LM to Read and Operate on the Web.}
In the initial stage, we mix the data constructed in Section~\ref{single_step} to equip the model with the ability to 
(1) comprehend the structure of web pages and the functions of various web components, and to (2) execute predefined operations on the current webpage, thus implementing simple user instructions. 

\vpara{To Make LM Learn to Plan \& Reason on the Web.}
During this stage, we continue to employ the constructed data in Section~\ref{multi_step} for training. We enable our model to decompose tasks into subtasks and execute subsequent steps based on the current webpage and the sequence of prior operations. 

After the above training, our model $M_{\text{SFT}}$ acquired essential capability in completing web browsing tasks and could independently execute operations based on user instructions.

\subsubsection{Step 2: Reinforcement Learning}\label{dpo}
Following previous training, $M_{\text{SFT}}$ has demonstrated some ability to operate the browser and infer the task. 
However, due to the distinctive nature of SFT training, $M_{\text{SFT}}$ attempts to mimic the inference process and operations but sometimes overlooks the webpage's state and preceding operation sequences, leading to hallucination. 
Consequently, we propose a self-sampling reinforcement learning to mitigate these operative illusions.

First, we use $M_{\text{SFT}}$ for $n$-fold sampling ($n$=20) on complex task operation samples in the training set. We combine the sampled output and golden answer to construct contrastive data with positive and negative pairs.
Subsequently, we retain samples based on the following criteria:
\begin{itemize}[leftmargin=*,itemsep=0pt,parsep=0.2em,topsep=0.2em,partopsep=0.0em]
    \item From all $n$ iterations of sampling, we select data where the model completed the tasks from 1 to $n$-1 times. 
If $M_{\text{SFT}}$ answered all iterations correctly, we consider it devoid of training value and incapable of providing practical negative examples. 
Conversely, If $M_{\text{SFT}}$ answered incorrectly across all iterations, we suspect issues with the data and exclude them, as the model cannot adequately fit these outliers during optimization.
\item  We retain different erroneous operations and remove duplicates to preserve distinct negative examples. 
\end{itemize}

After constructing contrastive data $D_{\text{Const.}}$, we employ the DPO\cite{rafailov2023direct} training approach to make $M_{\text{SFT}}$ learn from its mistakes and further enhance its capabilities.
During the training, we found that the direct use of DPO loss led to instability.
To mitigate this issue, we propose including SFT loss to stabilize the reinforcement learning process and increase the number of training steps while ensuring no loss of the original model's natural language and agent abilities, achieving a more robust model $M_{\text{DPO}}$:
\begin{equation}
\begin{array}{l}
\mathcal{L}_{DPO}(\pi_\theta;\pi_{\mathrm{ref}}) = \\
-\mathbb{E}_{(x,y_w,y_l)\thicksim\mathcal{D}}\left[\log\sigma\left(\beta\log\frac{\pi_\theta(y_w\mid x)}{\pi_{\mathrm{ref}}(y_w\mid x)}\right.\right.\left.\left.-\beta\log\frac{\pi_\theta(y_l\mid x)}{\pi_{\mathrm{ref}}(y_l\mid x)}\right)\right]
\end{array}
\end{equation}
\begin{equation}
\mathcal{L}_{SFT}(\pi_\theta;\pi_{\mathrm{ref}})=-\mathbb{E}_{(x,y_w,y_l)\thicksim\mathcal{D}}\left[\log \pi_\theta(y_w\mid x)\right]
\end{equation}
\begin{equation}
\mathcal{L}_{Total}=\lambda\cdot\mathcal{L}_{DPO}+\mathcal{L}_{SFT}
\end{equation}

\subsubsection{Step 3: Rejection Sampling Finetuning}\label{rft}
In the RFT (Rejection Sampling Finetuning) step, we aim to optimize for webpage environments in specific domains. 
RFT enables us to perform targeted training through substantial sampling from an existing model, selecting the accurate trajectories in instances lacking ones via reward signals. 
Our reward signals can be furnished either by the environment itself or through pre-designed reward models. 
Due to the network policy constraints inherent in real webpage environments, we conduct our experiments within sandbox environments furnished by MiniWob++ and WebArena.

For MiniWob++, we leverage the query generator in MiniWob++ to auto-generate multiple user queries for each task. We determine the number of generated queries for each task based on its difficulty.
Then, we employ $M_{\text{DPO}}$ to try to solve the queries. If a trace completes the task (as adjudged by the MiniWob++ environment), we consider this trace as a positive trace.

In the case of WebArena, to prevent overlap with the test set, we manually construct multiple distinctive user queries based on WebArena's templates. 
For each sample, we apply $M_{\text{DPO}}$ to perform 64 times of sampling.
Similarly, if our model completes the task at least once (adjudged by manually written rules), we deem the successful trace as a positive trace.

By utilizing the methods above, we constructed two distinct successful datasets, one from MiniWob++ and the other from WebArena. These comprise approximately 15k traces (66k steps) and 240 traces (2k steps), respectively, which are used for \model's individual finetuning on these two tasks.

\subsection{Benchmark: \benchmark}\label{autobench}
We segment the complex task operation dataset collected in Section~\ref{multi_step} for evaluation. 
\benchmark~is divided into two splits: in- and out-of-domain, which serve as bases for our performance assessment. 
The in-domain dataset represents training data collected from the same website, measuring the model's performance under familiar conditions.
In contrast, the out-of-domain dataset encompasses data collected from websites entirely excluded from our training set. It offers a unique opportunity to measure the model's generalizability and ability to adapt to unfamiliar environments.
We select 50 browsing traces for each split as our test data. These traces are scrutinized and filtered via human verification, ensuring a more reliable evaluation benchmark.

Drawing on the methodology presented in Mind2Web, we comprehensively evaluate each step involved in the operation. 
This allows us to assess the step and overall accuracy of the model's operations. Detailed results of this evaluation are available in Table~\ref{tab:autobench}.

\section{Experiments}
We establish a bilingual (Chinese-English) benchmark \benchmark and evaluate the abilities of publicly available agents.
We also conduct extensive experiments on numerous benchmarks to evaluate the performance of \model~in comparison to several baselines across various tasks involving navigating both English and Chinese websites.

\subsection{Main Results}
\vpara{\benchmark.}
As discussed in Section~\ref{autobench}, We divide the test set into four splits: Chinese, English, in-domain, and out-of-domain, for evaluation purposes.
We use the Step Success Rate (SSR) as our evaluation metric.
The results are in Table~\ref{tab:autobench}.

\begin{table}[t]
\caption{The performance on \benchmark.}
\label{tab:autobench}
\centering
\vspace{-4mm}
\renewcommand\tabcolsep{5pt}
\renewcommand\arraystretch{.9}
\resizebox{\columnwidth}{!}
{
\begin{tabular}{@{}lccccc@{}}
\toprule
& &\multicolumn{2}{c}{English} & \multicolumn{2}{c}{Chinese} \\
\cmidrule(lr){3-4} \cmidrule(lr){5-6}
Model & Size &\makecell{Cross-\\Task} & \makecell{Cross-\\Domain} & \makecell{Cross-\\Task} & \makecell{Cross-\\Domain} \\
\midrule
\chatgpt &N/A& 12.1 & 6.4 & 13.5 & 10.8 \\
\gptfour &N/A& 38.6 & 39.7 & 36.7 & 36.3 \\
Claude2 &N/A& 13.2 & 8.1 & 13.0 & 7.9 \\
LLaMA2 & 7B & 3.3 & 2.5 & - & - \\
LLaMA2 & 70B & 8.3 & 8.9 & - & - \\
Qwen &7B& 9.0 & 7.6 & 9.1 & 7.5 \\
\midrule
\model &6B& \textbf{64.8} & \textbf{58.6} & \textbf{65.4} & \textbf{61.8} \\
\bottomrule
\end{tabular}
}
\vspace{-3mm}
\end{table}

\vpara{Mind2Web.}
We use the settings from Mind2Web with SSR as our primary evaluation metric.
To compare the model fairly, we utilize the MindAct framework provided by Mind2Web to evaluate the model's performance.
The results are in Table~\ref{tab:mind2web}.
\begin{table}[t]
\caption{The performance on Mind2Web. \textmd{\dag~indicates that only top-10 candidates were used for this test, otherwise top-50 was used. * indicates model's finetuning on train set.}}
\label{tab:mind2web}
\centering
\vspace{-3mm}
\renewcommand\tabcolsep{5pt}
\renewcommand\arraystretch{.95}
\resizebox{\columnwidth}{!}
{
    \begin{tabular}{@{}lccccc@{}}
    \toprule
    Model & Size & \makecell{Cross-\\Task} & \makecell{Cross-\\Website} & \makecell{Cross-\\Domain} & Average \\ 
    \midrule
    \chatgpt &N/A& 17.4 & 16.2 & 18.6 & 17.4 \\
    \gptfour$^{\dag}$ &N/A& 36.2 & 30.1 & 26.4 & 30.9 \\
    Flan-T5-XL* &3B& 52.0 & 38.9 & 39.6 & 43.5 \\ 
    Html-T5-XL* &543B& \textbf{71.5} & \textbf{62.2} & \textbf{67.1} & \textbf{66.9} \\
    LLaMA2* & 7B& 52.7 & 47.1 & 50.3 & 50.1 \\
    LLaMA2* & 70B& 55.8 & 51.6 & 55.7 & 54.4 \\
    Qwen-VL* &9.6B & 12.6& 10.1& 8.0& 10.2\\
    SeeClick* &9.6B & 23.7& 18.8& 20.2& 20.9\\
    
    \midrule
    \model &6B& \underline{66.4} & \underline{56.4} & \underline{55.8} & \underline{59.5} \\
    \bottomrule
    \end{tabular}
}
\end{table}

\vpara{MiniWoB++ \& WebArena.}
For MiniWob++, following the experimental setup from WebAgent~\cite{gur2023real}, we test MiniWoB++ with 56 tasks by running 100 evaluation episodes per task to evaluate model capabilities.
For WebArena, we integrate our HTML parser module and action execution module into the WebArena environment to make it compatible with our system.
The results are in Table~\ref{tab:miniwobpp}.

\begin{table}[t]
\caption{The performance on MiniWoB++ and WebArena. \textmd{* indicates model's finetuning on train set.}}
\label{tab:miniwobpp}
\vspace{-3mm}
\centering
\resizebox{\columnwidth}{!}
{
\renewcommand\tabcolsep{14pt}
\renewcommand\arraystretch{.95}
    \begin{tabular}{@{}lccc@{}}
    \toprule
    Model& Size & MiniWoB++ &WebArena \\
    \midrule
    \chatgpt & N/A & 13.4 & 6.2\\
    \gptfour & N/A & 32.1 & 14.4 \\
    Text-Bison-001 & N/A &- & 5.1\\
    LLaMA2 & 7B & 42.8* & 1.2 \\
    LLaMA2 & 70B & 47.1* & 0.6 \\
    Html-T5-XL & 543B & 85.6* &-\\
    WebN-T5-XL & 3B & 48.4* &-\\ 
    Lemur & 70B & - & 5.3 \\
    \midrule
    \model & 6B &\textbf{89.3} &\textbf{18.2}\\
    \bottomrule
    \end{tabular}
}
\end{table}

\subsection{System Execution Efficiency}
\begin{table}
    \small
    \centering
    \caption{System Execution Efficiency}
    \label{tab:system_efficiency}
    \vspace{-4mm}
    \renewcommand\tabcolsep{3.5pt}
    \begin{tabular}{lcrrrrr}
    \toprule
    Action & count/tr & \multicolumn{1}{c}{Fetch} & \multicolumn{1}{c}{Parse} & \multicolumn{1}{c}{Predict} & \multicolumn{1}{c}{Execute} & \multicolumn{1}{c}{Loading} \\
    \midrule
    \verb|type_string|  & 1.00 & 362.00 &  28.88 &  2282.99 &  6.12 &  2082.61 \\
    \verb|click|        & 3.62 & 438.62 &  72.07 &  2252.10 & 28.10 &  2094.60 \\
    \verb|finish|       & 0.38 & 419.33 &  66.00 &  3054.16 &  6.00 &  2119.06 \\
    \verb|scroll_page|  & 0.38 & 475.33 &  88.67 &  3396.37 &  5.33 &  2033.35 \\
    Others              & 0.25 & 680.50 & 152.50 &  2666.08 & 10.00 &  2142.92 \\
    \midrule
    Average             & -    & 436.93 & 68.68  &  2407.34 & 20.36 &  2092.13 \\
    Percentage(\%)      & -    &   8.70 &  1.37  &    47.89 &  0.41 &    41.64 \\

    \bottomrule    
    \end{tabular}
\end{table}
Furthermore, since execution speed is critical to user experience, we conduct a series of performance experiments to evaluate the execution efficiency of each system component and identify areas that could be further optimized. The results of these experiments are presented in Table~\ref{tab:system_efficiency}.

\subsection{Ablation Study}
To evaluate the impact of different stages of data and training strategies on model performance enhancement, we conduct a comprehensive ablation study in Table~\ref{tab:ablation_study}.

\begin{table}[t]
\caption{Ablation study. \textmd{\benchmark and WebArena do not have a training set, while the RFT stage is only suitable for sampling in the environment, so we represent them by "-".}}
\label{tab:ablation_study}
\centering
\vspace{-4mm}
\renewcommand\tabcolsep{2pt}
\renewcommand\arraystretch{.95}
\resizebox{\columnwidth}{!}
{
\begin{tabular}{@{}lcccc@{}}
\toprule
Method & \benchmark & Mind2Web & MiniWob++ & WebArena \\ 
\midrule
\multicolumn{5}{c}{Training Data Ablation}\\  
\midrule
Only Train Set & - & 48.1 & 44.3 & - \\
+) Stage1 & 23.5 & 48.4 & 48.3 & 2.5 \\
+) Stage2 & 60.2 & 55.2 & 78.9 & 7.6 \\
+) Stage1+2 & 61.8 & 56.7 & 81.7 & 8.3 \\
\midrule
\multicolumn{5}{c}{Training Strategy Ablation}\\ 
\midrule
SFT & 61.8 & 56.7 & 81.7 & 8.3 \\
+) DPO & 62.7 & 59.5 & 80.8 & 8.5 \\
+) RFT & - & - & 89.3 & 18.2 \\
\midrule
\model & 62.7 & 59.5 & 89.3 & 18.2 \\
\bottomrule
\end{tabular}
}
\end{table}
\vpara{Training Data Ablation.}
We train and test only models that contain the original training set and incorporate simple and complex task data (see Section~\ref{data_preparation}) for training. This approach helps to qualitatively measure the impact of different datasets on the model.

The Complex Task dataset significantly improves model performance. We hypothesize that this is due to the complex data more closely aligning with real-world scenarios, thereby fundamentally transforming model performance.

The simple task dataset shows only a slight improvement when training alone. However, when training jointly with the complex task dataset, there is a significant improvement.
We find that training exclusively with complex task datasets leads to basic operational errors, suggesting that training with simple task datasets can effectively mitigate this problem.

\vpara{Training Strategy Ablation.}
We compare the results of SFT, DPO, and RFT-enhanced models and find that: 
(1) Compared to SFT, the DPO training facilitates model learning from its mistakes, further enhancing model performance.
(2) RFT enables our model to perform bootstrap enhancement in different domains. With practice comes proficiency, resulting in improvements within each domain.

\subsection{Case Study and Error Analysis}
To assess the effectiveness of our model, we conduct a series of case studies covering a range of web-based tasks, including everyday use, leisure and relaxation, and academic research, covering the typical range of web requirements. Our system achieves satisfactory results in most scenarios.

While our system performs commendably well on a variety of web-based tasks, it has limitations. We identify errors that occasionally occur during task execution, which can be broadly categorized into four types: hallucinations, poor graphical recognition, misinterpretation of task context, and pop-up interruptions. 
Table~\ref{tab:error_distribution} outlines the proportion of these errors observed during error analysis.
Although relatively infrequent, these errors are crucial in our ongoing efforts to refine and enhance the system's capabilities.

\begin{table}[t]
\centering
\caption{Error Distribution in Web Task Automation}
\vspace{-4mm}
\label{tab:error_distribution}
\renewcommand\tabcolsep{10pt}
\renewcommand\arraystretch{.95}
\begin{tabular}{lc}
\toprule
Error Type & Proportion \\
\midrule
Hallucinations & 44\% \\
Poor Graphical Recognition & 28\% \\
Misinterpretation of Task Context & 20\% \\
Pop-Up Interruption & 8\% \\
\bottomrule
\end{tabular}
\end{table}

\section{Future Direction}
\subsection{Multimodal Input}
While HTML input has produced satisfactory results in many scenarios, our system falters when confronted with advanced web applications such as maps, animations, and video browsing. In our analysis, the strength of image input lies in its indispensable role in interpreting images, icons, and special effects. 
However, compared to text input, image input presents additional challenges in understanding numerals and extensive web text. Consequently, we consider that a multimodal system, integrating HTML and webpage screenshots, combines the advantages of both modalities, substantially enhancing the model's capability in web browsing tasks.

\subsection{Reasoning and Self-check Techniques}
The system's efficiency and success rate in web browsing may decrease when dealing with unfamiliar websites or those with unique operating logic. To mitigate this issue, an exciting avenue for exploration is the development of novel reasoning strategies distinct from the Chain-of-Thought approach, enabling the model to make better-informed decisions based on previous browsing experiences, thereby improving the success rate and efficiency of web browsing. Moreover, due to unstable internet connections and other factors, the stability of a real web environment is not guaranteed. Thus, self-check mechanisms within the web browsing agent system, including confirming the current state and verifying the intended operation's effect, could significantly improve the system's robustness and effectiveness.

\subsection{Mobile Application}
The mobile platform is another promising application scenario with massive potential. Compared to the web platform, it presents its challenges and opportunities. For example, due to their screen size, mobile devices display fewer elements within the viewport, simplifying the page XML. Furthermore, the operation logic on mobile platforms is generally more straightforward than on web platforms. 
However, mobile operation space includes more complex actions such as gestures, and mobile platforms face more system security restrictions, imposing additional constraints on software development.

\section{Conclusion}
In this work, we present \model, an advanced language model-based agent exhibiting robust performance in various autonomous web navigation benchmarks. 
Our model addresses extant LLM limitations and simplifies webpages by effectively controlling HTML text length and handling the web's open-domain nature. 
We strategically employ curriculum learning, reinforcement learning, and rejection sampling finetuning to enhance webpage comprehension and browser operation learning. 
We also introduce a unique bilingual web browsing benchmark--- that lays a solid foundation for future research. Our findings represent significant progress in utilizing LLMs for intelligent agent tasks.


\begin{acks}
This work is supported by Natural Science Foundation of China (NSFC) 62276148  and 62425601, the New Cornerstone Science Foundation through the XPLORER PRIZE and  Tsinghua University (Department of Computer Science and Technology) -Siemens Ltd., China Joint Research Center for Industrial Intelligence and Internet of Things (JCIIOT).
\end{acks}

\clearpage
\bibliographystyle{ACM-Reference-Format}
\balance
\bibliography{sample-base}

\clearpage
\appendix
\section{Implementation Details of \model}
During the SFT phase, we set the learning rate to 1e-5 with a batch size of 32. 
In the DPO stage, we sample the complex task dataset 20 times. After the filtering process, we build a contractional dataset of approximately 13k. We set the learning rate for the DPO to 1e-6, the batch size to 64, and the $\beta$ parameter to 0.15. We add the SFT loss, weighted by a factor of 0.8.
During the RFT stage, we collect samples from two diverse environments, MiniWoB++ and WebArena, resulting in successful datasets of approximately 66k and 2k, respectively, which underwent finetuning. The learning rate set for this stage was 1e-5, and the batch size was 32.
\section{Input Prompt}\label{appendix:input_prompt}
Below is our input prompt for \model:
\lstset{
    backgroundcolor=\color[RGB]{245,245,244},
    breaklines=true,
    basicstyle=\ttfamily\small
}\begin{lstlisting}
<html> {html_content} </html>

You are a helpful assistant that can assist with web navigation tasks.
You are given a simplified html webpage and a task description.
Your goal is to complete the task. You can use the provided functions below to interact with the current webpage.

#Provided functions:
def click(element_id: str) -> None:
    """
    Click on the element with the specified id.

    Args:
       element_id: The id of the element.
    """

... (Other function definitions)

#Previous commands: {previous_commands}

#Window tabs: {exist_window_tabs_with_pointer_to_current_tab}

#Current viewport (pages): {current_position} / {max_size}

#Task: {task_description}

You should output one command to interact to the currrent webpage.
You should add a brief comment to your command to explain your reasoning and thinking process.\end{lstlisting}
\section{Data Construction Prompt}

Data construction prompt for task and trace intent:
\lstset{
    backgroundcolor=\color[RGB]{245,245,244},
    breaklines=true,
    basicstyle=\ttfamily\small
}\begin{lstlisting}
HTML:
{html_content}
I want you to act as a task generator that can help generate Task-Operation pairs.
Based on the above HTML webpage, I will give you a specified operation. Your goal is to come up with a ONE-STEP task that the specified operation can solve.
Your answer SHOULD be in the following format:

Task: {Generated one-step task}

Operation: {The right operation to solve the task}

Intention: {The intention and thinking in your operation}

NOTICE: 
1. Your generated task should not be too SIMPLE, NAIVE
2. You can only do \#type\# on <input> and <textarea>
\end{lstlisting}
\lstset{
    backgroundcolor=\color[RGB]{245,245,244},
    breaklines=true,
    basicstyle=\ttfamily\small
}\begin{lstlisting}
User's overall task: {task_description}

User's actions: {annotated_action_trace}

Based on this information, deduce the intent behind each of the user's actions. Your response should be structured as follows:
Intent of Step 1: [Describe the intent of the user's first action from the user's first-person perspective]
Intent of Step 2: [Describe the intent of the user's second action from the user's first-person perspective]
... and so on.
Note: Your response should have the same number of lines as the number of user actions. The number of user actions in this task is {number_of_steps_in_action}.

\end{lstlisting}
\section{Annotation Details}
The annotation process was performed by 20 annotators for one month using the Google Chrome browser with our plugin installed to record their actions on assigned websites. The annotators first visited the target websites and checked whether the website descriptions matched the actual tasks. They then evaluated the tasks for clarity, relevance, achievability, complexity, and subjectivity, skipping those that didn't meet the criteria.
They carefully recorded each step during a task, including any login or captcha steps. 
For tasks that required an answer, the annotators manually edited the responses.
If a task was not doable, they could modify its description or abandon it.
\section{Full Results of MiniWob++}
Table~\ref{tab:per_task_performance_on_miniwobpp} is the per-task average success rate on 56 tasks from MiniWoB++.
\begin{table*}
\caption{PER-TASK PERFORMANCE ON MINIWOB++}
\label{tab:per_task_performance_on_miniwobpp}
\small
\renewcommand\tabcolsep{3pt}
\renewcommand\arraystretch{1}
\begin{tabular}{@{}l|p{2cm}<{\centering}|p{2cm}<{\centering}p{2cm}<{\centering}|p{2cm}<{\centering}p{2cm}<{\centering}@{}}
\toprule
Task & \model & HTML-T5-XL & WebN-T5-XL & \gptfour &\chatgpt \\ 
\midrule
book-flight & 0.50 & 0.99 & 0.48 & 0.00 & 0.00\\
choose-date & 1.00 & 0.16 & 0.08 & 0.00 & 0.00\\
choose-date-easy & 1.00 & 1.00 & 1.00 & 0.00 & 0.00\\
choose-date-medium & 1.00 & 0.56 & 0.07 & 0.00 & 0.00\\
choose-list & 0.15 & 0.22 & 0.16 & 0.00 & 0.00\\
click-button & 1.00 & 1.00 & 1.00 & 0.67 & 1.00\\
click-button-sequence & 1.00 & 1.00 & 1.00 & 0.33 & 0.00\\
click-checkboxes  & 1.00 & 1.00 & 0.22 & 0.33 & 0.00\\
click-checkboxes-large & 0.83 & 0.90 & 0.54 & 0.00 & 0.00\\
click-checkboxes-soft& 0.37 & 0.99 & 0.08 & 0.00 & 0.00\\
click-checkboxes-transfer & 1.00 & 1.00 & 0.63 & 1.00 & 0.00\\
click-collapsible & 1.00 & 1.00 & 0.26 & 0.00 & 0.00\\
click-collapsible-2 & 0.76 & 0.93 & 0.27 & 0.00 & 0.00\\
click-color & 0.74 & 1.00 & 0.34 & 0.67 & 0.00\\
click-dialog & 1.00 & 1.00 & 1.00 & 0.33 & 0.00\\
click-dialog-2& 1.00 & 0.74 & 1.00 & 0.67 & 0.67\\
click-link & 1.00 & 1.00 & 0.99 & 0.33 & 0.33\\
click-menu & 1.00 & 0.37 & 0.41 & 0.00 & 0.50\\
click-option & 1.00 & 1.00 & 0.87 & 0.67 & 0.00\\
click-pie & 1.00 & 0.96 & 0.51 & 0.67 & 1.00\\
click-scroll-list & 0.57 & 0.99 & 0.98 & 0.00 & 0.00\\
click-shades & 1.00 & 0.00 & 0.00 & 0.00 & 0.00\\
click-shape  & 0.64 & 0.79 & 0.24 & 0.00 & 0.00\\
click-tab & 1.00 & 1.00 & 0.57 &0.00 & 0.67\\
click-tab-2 & 1.00 & 0.94 & 0.57 &0.00 & 0.00\\
click-tab-2-hard & 1.00 & 0.88 & 0.12 & 0.33 & 0.00\\
click-test & 1.00 & 1.00 & 1.00 & 1.00 & 0.00\\
click-test-2 & 0.93 & 1.00 & 1.00 & 1.00 & 1.00\\
click-widget & 1.00 & 1.00 & 1.00 & 1.00 & 0.00\\
count-shape & 0.65 & 0.67 & 0.64 & 0.00 & 0.00\\
email-inbox & 1.00 & 1.00 & 0.38 & 0.00 & 0.33\\
email-inbox-forward-nl & 1.00 & 1.00 & 0.33 & 0.00 & 0.00\\
email-inbox-forward-nl-turk & 1.00 & 1.00 & 0.23 & 0.00 & 0.00\\
email-inbox-nl-turk & 1.00 & 0.99 & 0.20 & 0.67 & 0.00\\
enter-date & 1.00 & 1.00 & 0.89 & 0.66 & 0.00\\
enter-password & 1.00 & 1.00 & 0.72 & 0.67 &0.00\\
enter-text & 1.00 & 1.00 & 0.89 & 0.67 &0.00\\
enter-text-dynamic & 1.00 & 1.00 & 1.00 & 0.00 &0.00\\
enter-time & 0.00 & 1.00 & 0.00 & 0.00 & 0.00\\
focus-text & 1.00 & 1.00 & 1.00 & 0.00 & 0.00\\
focus-text-2 & 1.00 & 1.00 & 1.00 & 0.00 & 1.00\\
grid-coordinate & 1.00 & 1.00 & 1.00 & 1.00 & 0.33\\
guess-number & 1.00 & 0.13 & 0.00 & 0.00 & 0.00\\
identify-shape & 1.00 & 1.00 & 0.88 & 0.67 & 0.00\\
login-user & 1.00 & 1.00 & 0.82 & 0.33 & 0.00\\
login-user-popup & 0.63 & 1.00 & 0.72 & 0.33 & 0.00\\
multi-layouts & 1.00 & 1.00 & 0.83 & 0.33 & 0.00\\
multi-orderings & 1.00 & 1.00 & 0.88 & 0.67 & 0.00\\
navigate-tree & 1.00 & 0.99 & 0.91 & 0.33 & 0.00\\
search-engine & 1.00 & 0.93 & 0.34 & 0.67 & 0.00\\
social-media & 1.00 & 0.99 & 0.20 & 0.33 & 0.00\\
social-media-all & 0.90 & 0.31 & 0.21 & 0.00 & 0.00\\
social-media-some & 0.76 & 0.89 & 0.42 & 0.00 & 0.00\\
tic-tac-toe & 0.74 & 0.57 & 0.48 & 0.00 & 0.00\\
use-autocomplete & 0.85 & 0.97 & 0.02 & 1.00 & 0.67\\
use-spinner & 1.00 & 0.07 & 0.07 & 0.00 &0.00\\
\midrule
Average & \textbf{0.893} & 0.856 & 0.484 & 0.321 & 0.134 \\
\bottomrule
\end{tabular}
\end{table*}

\end{document}